\documentclass{article}
\usepackage{mathtools}
\usepackage{xfrac}
\usepackage{amsfonts}
\usepackage{amsthm}
\usepackage{subfig}
\usepackage{hyperref}
\usepackage{xcolor}

\usepackage{times}
\usepackage{float}

\usepackage[square,numbers]{natbib}
\usepackage[final]{bdl_2018}

\usepackage{bbm}
\usepackage{algorithm, algorithmic}
\usepackage{wrapfig}
\usepackage{lipsum}

\allowdisplaybreaks

\title{Uncertainty propagation in neural networks for sparse coding}

\author{
Danil Kuzin$^{\dagger}$, Olga Isupova$^{\star}$, Lyudmila Mihaylova$^{\dagger}$\\
$^{\dagger}$Department of Automatic Control
and System Engineering,
University of Sheffield, UK \\ 
$^{\star}$Department of Engineering Science,
University of Oxford, UK \\
\texttt{\{dkuzin1,l.s.mihaylova\}@sheffield.ac.uk,olga.isupova@eng.ox.ac.uk}
}

\begin{document}
\maketitle
\section{Introduction}
  The idea of Bayesian learning in neural networks (NNs)~\cite{neal2012bayesian} has recently gained an attention with the development of distributed approximate inference techniques~\cite{li2015stochastic, hoffman2013stochastic} and general boost in popularity of deep learning. Recently several techniques~\cite{ranganath2015deep, gal2016dropout} have been proposed to handle specific types of NNs with efficient Bayesian inference. For example, feed-forward networks with the rectified linear unit nonlinearity~\cite{hernandez2015probabilistic}, networks with discrete distributions~\cite{soudry2014expectation}, recurrent networks~\cite{mcdermott2017bayesian}.

  In this paper, we consider the area of sparse coding. The sparse coding problem can be viewed as a linear regression problem with the additional assumption that the majority of the basis representation coefficients should be zeros. This sparsity assumption may be represented as $l1$ penalty~\cite{tibshirani1996regression}, or, in Bayesian interpretation, as a prior that has a sharp peak at zero~\cite{tipping2001sparse}. One of the modern approaches for sparse coding utilises NNs with the soft-thresholding nonlinearity~\cite{gregor2010learning, sprechmann2015learning}. Sparse coding is widely used in different applications, such as compressive sensing~\cite{candes2008introduction}, image and video processing~\cite{mairal2014sparse, wang2015deep}, neuroscience~\cite{baillet1997bayesian, jas2017learning}.

  A novel method to propagate uncertainty through the soft-thresholding nonlinearity is proposed in this paper. At every layer the current distribution of the target vector is represented as a spike and slab distribution~\cite{mitchell1988bayesian}, which represents the probabilities of each variable being zero, or Gaussian-distributed. Using the proposed method of uncertainty propagation, the gradients of the logarithms of normalisation constants are derived, that can be used to update a weight distribution. A novel Bayesian NN for sparse coding is designed utilising both the proposed method of uncertainty propagation and Bayesian inference algorithm.

  The main contributions of this paper are: (\textit{i}) for the first time a method for uncertainty propagation through the soft-thresholding nonlinearity is proposed for a Bayesian NN; (\textit{ii}) an efficient posterior inference algorithm for weights and outputs of NNs with the soft-thresholding nonlinearity is developed; (\textit{iii}) a novel Bayesian NN for sparse coding is designed.

 The rest of the paper is organised as follows. A NN approach for sparse coding is described in Section~\ref{subsec:nn_sc}. The Bayesian formulation is introduced in Section~\ref{subsec:bayesian_lista}. Section~\ref{sec:experiments} provides the experimental results. The proposed forward uncertainty propagation and probabilistic backpropagation methods are given in Appendices~\ref{sec:fprop} and~\ref{sec:backpropagation}. 
\section{Neural networks for sparse coding}
  \label{sec:bayesian_lista}
  This section presents background knowledge about networks for sparse coding and then describes the novel Bayesian neural network. 
\subsection{Frequentist neural networks}  
\label{subsec:nn_sc}
  The NN approach to sparse coding is based on earlier Iterative Shrinkage and Thresholding Algorithm (ISTA) \cite{daubechies2004iterative}. It addresses the sparse coding problem as the linear regression problem with the $l1$ penalty that promotes sparsity. For the linear regression model with observations $\mathbf{y} \in \mathbb{R}^K$, the design matrix $\mathbf{X} \in \mathbb{R}^{K \times D}$, and the sparse unknown vector of weights $\boldsymbol\beta \in \mathbb{R}^D$, ISTA minimises
  \begin{equation}
  \label{eq:regression_problem}
  ||\mathbf{X}\boldsymbol\beta - \mathbf{y}||_2^2 + \alpha ||\boldsymbol\beta||_1 \, \text{w.r.t.} \, \boldsymbol\beta,
  \end{equation}
where $\alpha$ is a regularisation parameter.
 
 \begin{wrapfigure}[13]{r}{0.5\textwidth}  
 \vspace{-25pt}
\begin{minipage}{0.5\textwidth}
  \begin{algorithm}[H]
    \caption{LISTA forward propagation}
\label{alg:lista}   
    \begin{algorithmic}[1]
      \REQUIRE observations $\mathbf{y}$, weights $\mathbf{W}, \mathbf{S}$, number of layers $L$
      \STATE Dense layer $\mathbf{b} \gets \mathbf{W}\mathbf{y}$ \label{eq:first_layer}
      \STATE Soft-thresholding function~$\widehat{\boldsymbol\beta}_0 \gets h_\lambda(\mathbf{b})$ \label{eq:thr_first}
      \FOR{$l=1$ \TO $L$}
        \STATE Dense layer $\mathbf{c}_l \gets \mathbf{b} + \mathbf{S}\widehat{\boldsymbol\beta}_{l-1}$ \label{eq:l_dense_layer}
        \STATE Soft-thresholding function $\widehat{\boldsymbol\beta}_{l} \gets h_\lambda(\mathbf{c}_l)$ \label{eq:l_thr}
      \ENDFOR
      \RETURN $\widehat{\boldsymbol\beta} \gets \widehat{\boldsymbol\beta}_{L}$
    \end{algorithmic}
  \end{algorithm}
\end{minipage}
\end{wrapfigure}  
At every iteration $l$, ISTA obtains the new estimate $\widehat{\boldsymbol\beta}_l$ of the target vector $\boldsymbol\beta$ as the linear transformation $\mathbf{b} = \mathbf{W}\mathbf{y} + \mathbf{S}\widehat{\boldsymbol\beta}_{l-1}$ propagated through the soft-thresholding function 
  \begin{equation}
  h_\lambda(\mathbf{b}) = \text{sgn}(\mathbf{b}) \max(|\mathbf{b}| - \lambda, 0),
  \end{equation}
  where $\lambda$ is a shrinkage parameter.
  In ISTA, weights $\mathbf{W}$ and $\mathbf{S}$ of the linear transformation are assumed fixed.
   
In contrast to ISTA, Learned ISTA (LISTA) \cite{gregor2010learning} learns the values of matrices $\mathbf{W}$ and $\mathbf{S}$ based on a set of pairs $\{\mathbf{Y}, \mathbf{B}\}=\{\mathbf{y}^{(n)}, \boldsymbol\beta^{(n)}\}_{n=1}^N$, where $N$ is the number of these pairs. To achieve this, ISTA is limited with the fixed amount of iterations $L$ and interpreted as a recurrent NN: every iteration $l$ of ISTA corresponds to the layer $l$ of LISTA. A vector $\widehat{\boldsymbol\beta}$ for an observation~$\mathbf{y}$ is predicted by Algorithm~\ref{alg:lista}.


  \subsection{BayesLISTA}
  \label{subsec:bayesian_lista}
  This section introduces the proposed Bayesian version of LISTA (BayesLISTA). The prior distributions are imposed on the unknown weights
  \begin{equation}
  \label{eq:ws}
  p(\mathbf{W}) = \prod_{d=1}^D\prod_{k=1}^K \mathcal{N}(w_{ij} ; 0, \eta^{-1}), \quad
  p(\mathbf{S}) = \prod_{d'=1}^D\prod_{d''=1}^D \mathcal{N}(s_{d'd''} ; 0, \eta^{-1}),
  \end{equation}
  where $\eta$ is the precision of the Gaussian distribution.
  
  For every layer $l$ of BayesLISTA, $\widehat{\boldsymbol\beta}_{l}$ is assumed to have the spike and slab distribution with the spike probability $\boldsymbol\omega$, the slab mean $\mathbf{m}$, and the slab variance $\mathbf{v}$
  \begin{equation}
  [\widehat{\boldsymbol\beta}_{l}]_d \sim \omega_d \delta_0 + (1 - \omega_d)\mathcal{N}(m_d, v_d),
  \end{equation}
  where $\delta_0$ is the delta-function that represents a spike, $[\cdot]_d$ denotes the $d$-th component of a vector. In appendix we show that the output of the next layer $\widehat{\boldsymbol\beta}_{l+1}$ can be approximated with the spike and slab distribution and, therefore, the output of the BayesLISTA network $\widehat{\boldsymbol\beta}$ has the spike and slab distribution.
  
  To introduce the uncertainty of predictions, we assume that the true $\boldsymbol\beta$ is an output $f(\mathbf{y} ; \mathbf{S}, \mathbf{W}, \lambda)$ of the BayesLISTA network corrupted by the additive Gaussian zero-mean noise with the precision~$\gamma$. Then the likelihood of $\mathbf{B}$ is defined as
  \begin{equation}
  \label{eq:likelihood}
  p(\mathbf{B}| \mathbf{Y}, \mathbf{W}, \mathbf{S}, \gamma, \lambda)
  = \prod_{n=1}^N\prod_{d=1}^D\mathcal{N}\left(\beta_d^{(n)}; [f(\mathbf{y} ; \mathbf{S}, \mathbf{W}, \lambda)]_d, \gamma^{-1}\right)
  \end{equation}
  Gamma prior distributions with parameters~$a^{\cdot}$ and~$b^{\cdot}$ are specified on the introduced Gaussian precisions
  \begin{equation}
  \label{eq:gamma_eta}
  p(\gamma) = \text{Gam}\left(\gamma; a^{\gamma}, b^{\gamma}\right), \qquad
  p(\eta) = \text{Gam}\left(\eta; 	a^{\eta}, b^{\eta}\right)
  \end{equation}
  
  The posterior distribution is then
  \begin{equation}
  \label{eq:posterior}
  p(\mathbf{W}, \mathbf{S}, \gamma, \eta | \mathbf{B}, \mathbf{Y}, \lambda)
  = \frac{p(\mathbf{B} | \mathbf{Y}, \mathbf{W},  \mathbf{S}, \gamma, \lambda) p(\mathbf{W} | \eta )p(\mathbf{S} | \eta) p(\eta) p(\gamma)}{p(\mathbf{B} | \mathbf{Y}, \lambda)}
  \end{equation}
  The shrinkage parameter $\lambda$ is a hyperparameter of the model.
  
In the appendix we describe modification of LISTA forward propagation (Algorithm~\ref{alg:lista}) to include probability distributions of the random variables introduced in this section and also an efficient Bayesian inference algorithm.
  
  \section{Experiments}
  \label{sec:experiments}

   \begin{figure}[!t]
  \centering
  \subfloat[Synthetic for different $L$]{\includegraphics[width=0.28\textwidth]{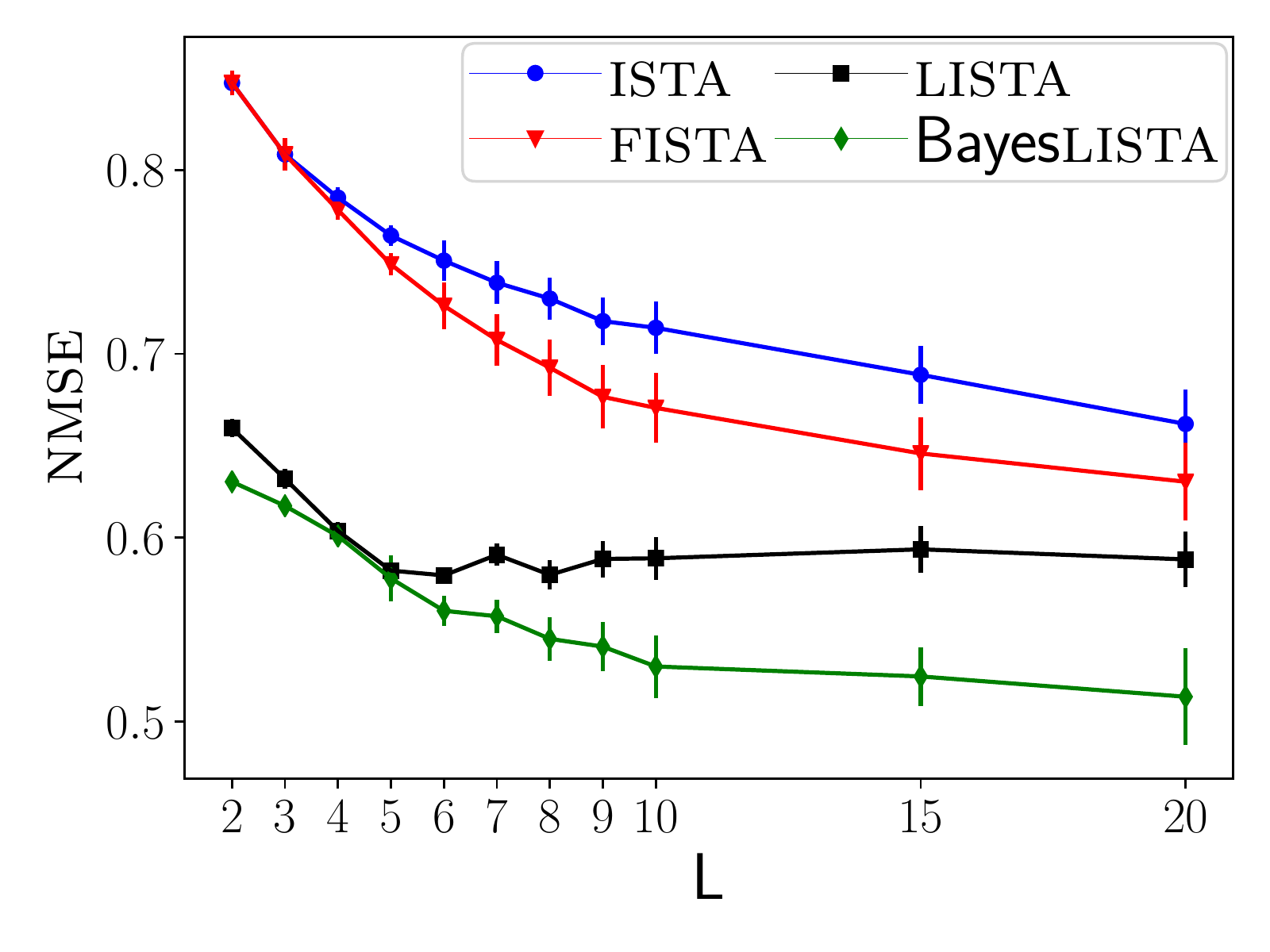}
  \label{fig:nmse_n_layers_synthetic}}
  \hfil
  \subfloat[Synthetic for different $K$]{\includegraphics[width=0.28\textwidth]{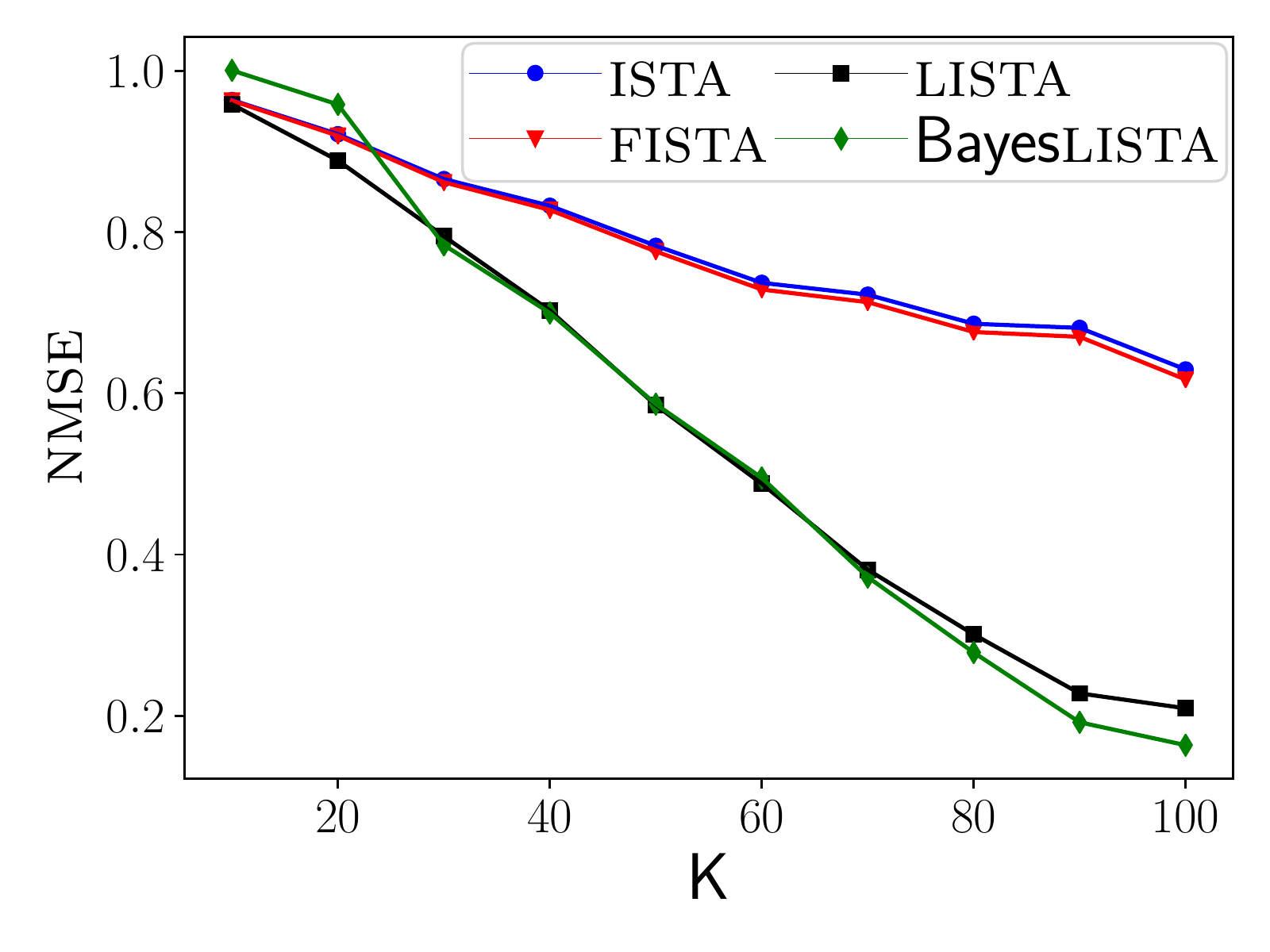}
  \label{fig:nmse_undersampling_synthetic}}
  \hfil
  \subfloat[Active learning example]{\includegraphics[width=0.28\textwidth]{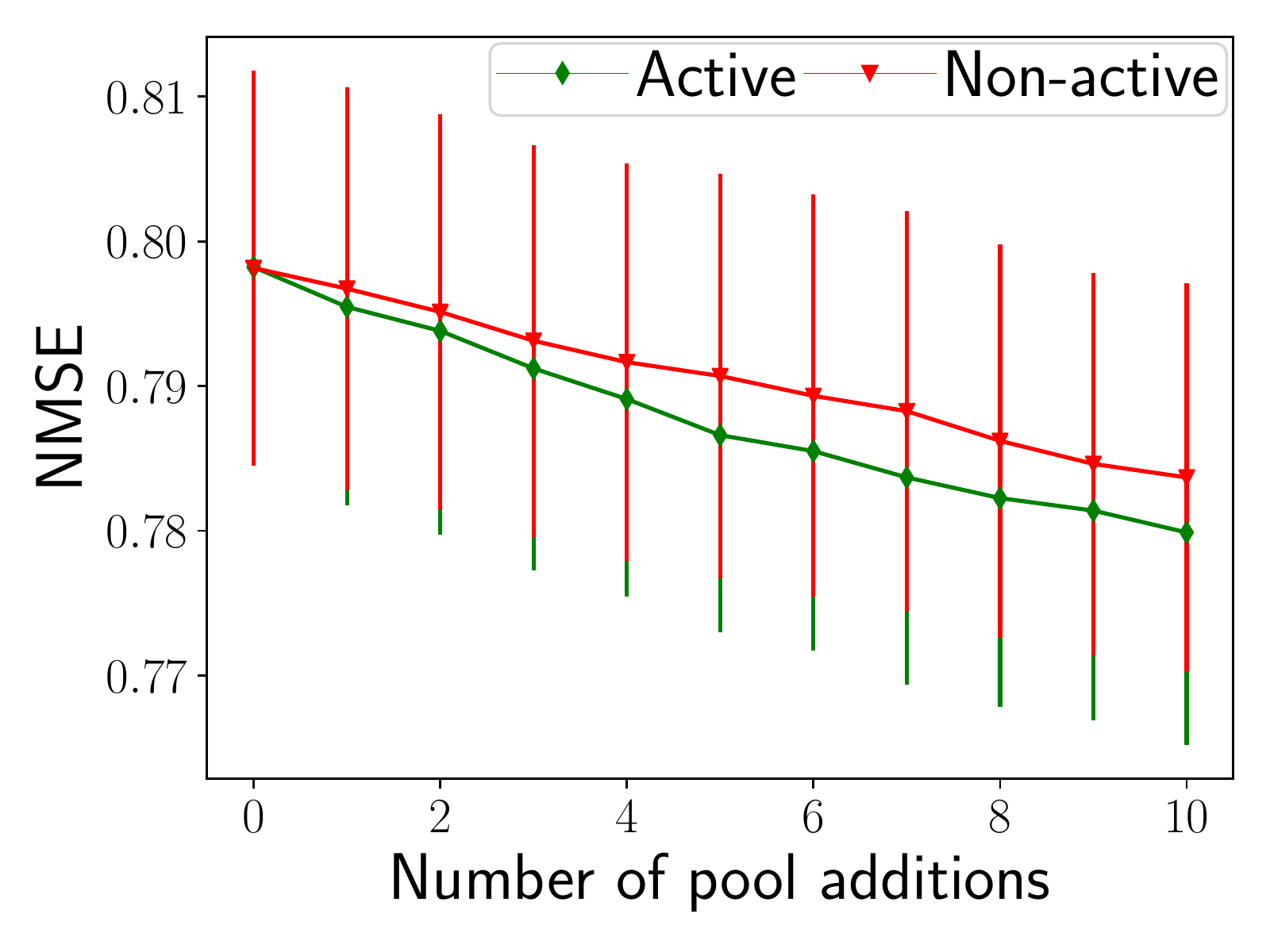}
  \label{fig:active_learning_mnist}}
  \caption{NMSE results. The synthetic data results for different number of layers~\protect\subref{fig:nmse_n_layers_synthetic} and for different sizes of observations~\protect\subref{fig:nmse_undersampling_synthetic}. The active learning example results on the MNIST data~\protect\subref{fig:active_learning_mnist}.}
  \label{fig:number_of_layers_synthetic}
  \end{figure}  
  
  Proposed BayesLISTA is evaluated on sparse coding problems and compared with LISTA~\cite{gregor2010learning}, ISTA \cite{daubechies2004iterative} and Fast ISTA (FISTA) \cite{beck2009fast}. The number of iterations in ISTA and FISTA  and the number of layers in NNs is $L$. For quantitative comparison the normalised mean square error (NMSE) is used. 
  
  \subsection{Predictive performance on synthetic data}
  First, performance is analysed on synthetic data. We generate $N_\text{train}=1000$ and $N_{\text{test}} = 100$ sparse vectors $\boldsymbol\beta^{(n)}$ of size $D = 100$  from the spike and slab distribution with the truncated slab: each component $\beta^{(n)}_{d}$ is zero with the probability~$0.8$ or is sampled from the standard Gaussian distribution without interval $(-0.1, 0.1)$ with the probability~$0.2$. The design matrix $\mathbf{X}$ is random Gaussian.  The observations $\mathbf{y}^{(n)}$ are generated as in~(\ref{eq:regression_problem}) with the zero-mean Gaussian noise with the standard deviation $0.5$. The shrinkage parameter is set to~$\lambda = 0.1$. The algorithms are trained on the training data of size $N_\text{train}$ and evaluated on the test data of size $N_{\text{test}}$.
  
  In \figurename~\ref{fig:nmse_n_layers_synthetic} NMSE for different number of layers (or iterations) $L$ is presented. The observation size is set to $K=50$. BayesLISTA outperforms competitors.
   \figurename~\ref{fig:nmse_undersampling_synthetic} gives NMSE for different observation sizes $K$. The number of layers (iterations) is set as $L=4$. In the previous experiment, Bayesian and classic LISTA show similar results with this number of layers. \figurename~\ref{fig:nmse_undersampling_synthetic} confirms this competitive behaviour between two LISTAs. ISTA and FISTA underperform the NNs. 
  
  \subsection{Active learning}
 To demonstrate a potential scenario that can benefit from uncertainty estimates of BayesLISTA, we consider the active learning example~\cite{settles.tr09}. The active learning area researches ways to select new training subsets to reduce the total number of required supervision. One of the popular approaches in active learning is uncertainty sampling, when the data with the least certain predictions is chosen for labelling. We use a variance of the spike and slab distributed prediction as a measure of uncertainty.
  
 The MNIST dataset~\cite{lecun1998gradient} is utilised. The dataset contains images of handwritten digits of size $28 \times 28 = 784$. The design matrix $\mathbf{X}$ is standard random Gaussian. Observations are generated as $\mathbf{y} = \mathbf{X}\boldsymbol\beta$, where $\boldsymbol\beta \in \mathbb{R}^{784}$ are flattened images. The shrinkage parameter $\lambda$ is~$0.1$, the observation size $K$ is $100$.

 We use the training data of size $50$, the pool data of size $500$,  and the test data of size $100$. The algorithm learns on the training data and it is evaluated on the test data. To actively collect a next data point from the pool, the algorithm is used to predict a point with the highest uncertainty. The selected point is moved from the pool to the training data and the algorithms learns on the updated training data. Overall, $10$ pool additions are performed. After every addition the performance is measured on the test data. We compare the active approach of adding new points from the pool with the random approach that picks a new data point from the pool at random. The procedure is repeated for $20$ times.
  
  Figure \ref{fig:active_learning_mnist} demonstrates performance of the active and non-active methods of updates with BayesLISTA. The active approach with uncertainty sampling steadily demonstrates better results. This means the posterior distribution learnt by BayesLISTA is an adequate estimate of the true posterior.
  
 Appendix~\ref{app:mnist} provides additional results on predictive performance on the MNIST data.

{\small
\bibliography{bibliography}
\bibliographystyle{unsrtnat}
}

\appendix
  \section{Appendix: Uncertainty propagation through soft-thresholding}
  \label{sec:fprop}
  This section describes modification of LISTA forward propagation (Algorithm~\ref{alg:lista}) to include probability distributions of the random variables introduced in section~\ref{subsec:bayesian_lista}. 
  
  \subsection*{Initialisation}
  \label{subsec:forward_init}
  At step \ref{eq:first_layer} of LISTA (Algorithm~\ref{alg:lista}) the matrix $\mathbf{W}$ consists of Gaussian-distributed components $w_{dk} \sim \mathcal{N}(m^w_{dk}, v^w_{dk})$, and $\mathbf{y}$ is a deterministic vector. Then the output $\mathbf{b}$ is a vector of Gaussian-distributed components $b_d \sim \mathcal{N}(m^b_d, v^b_d)$, where $m^b_d = \sum_{k=1}^Ky_k m^w_{dk}$, and $v^b_d = \sum_{k=1}^Ky_k^2v^w_{dk}$.
  
  At step~\ref{eq:thr_first} of LISTA (Algorithm~\ref{alg:lista}) the Gaussian vector $\mathbf{b}$ is taken as an input of the soft-thresholding function. When a Gaussian random variable $x \sim \mathcal{N}(x; m, v)$ is propagated through the soft-thresholding function $x^* = h_{\lambda}(x)$, the probability mass of the resulting random variable $x^*$ is split into two parts. The values of $x$ from the interval $[-\lambda, \lambda]$ are converted to~$0$ by the soft-thresholding operator. Therefore, the probability mass of the original distribution that lies in $[-\lambda, \lambda]$ is squeezed into the probability of $x^*$ being zero. The values of $x$ from outside of the $[-\lambda, \lambda]$ interval are shifted towards~$0$. The distribution of $x^* \neq 0$ then represents the tails of the original Gaussian distribution. The distribution of $x^*$ can be then parametrised by the probability of being zero, $\omega^*$, the mean $m^*$ and the variance $v^*$ of the truncated Gaussian distribution.
 Therefore, we approximate the distribution of $\widehat{\boldsymbol\beta}_0$ at step~\ref{eq:thr_first} with a spike and slab distribution with parameters: the spike probability $\omega^*$, the slab mean $m^*$ and variance $v^*$.
  
  \subsection*{Main layers}
  At step \ref{eq:l_dense_layer} of LISTA (Algorithm~\ref{alg:lista}) the vector $\mathbf{b}$ and matrix $\mathbf{S}$ consist of Gaussian components: $b_d \sim \mathcal{N}(m^b_d, v^b_d)$, $s_{d'd''} \sim \mathcal{N}(m^s_{d'd''}, v^s_{d'd''})$, and $\widehat{\boldsymbol\beta}_{l-1}$ is a vector of the spike and slab random variables: $[\widehat{\beta}_{l-1}]_d \sim \omega_d \delta_0 + (1 - \omega_d)\mathcal{N}(m_d, v_d)$. 
  
 It can be shown that the expected value and variance of a spike and slab distributed variable $\xi$ with the probability of spike $\omega$, the slab mean $m$ and slab variance $v$ are:  
   \begin{equation}
   \label{eq:spsl_moments}
  \mathbb{E}\xi = (1-\omega)m, \qquad \operatorname{Var}\xi = (1-\omega)(v + \omega m^2).
  \end{equation}
  
 It can also be shown that if components of the matrix $\mathbf{S}$ and vector $\widehat{\boldsymbol\beta}_{l-1}$ are mutually independent then the components $[ \mathbf{e}_l ]_d$ of their product $\mathbf{e}_l = \mathbf{S} \widehat{\boldsymbol\beta}_{l-1}$ have the marginal mean and variances:
  \begin{subequations}
  \label{eq:e_moments}
  \begin{align}
  m^e_{d} \stackrel{\text{def}}{=} &\mathbb{E}[ \mathbf{e}_l ]_d = \sum_{d'=1}^D m^s_{dd'}(1-\omega_{d'})m_{d'}, \\
  v^e_{d} \stackrel{\text{def}}{=} &\operatorname{Var}[ \mathbf{e}_l ]_d = \sum_{d'=1}^D [(m^s_{dd'})^2(1-\omega_{d'})^2v_{d'}
   + (1-\omega_{d'})^2(m_{d'})^2v^s_{dd'} + v^s_{dd'}(1-\omega_{d'})^2v_{d'}].
   \end{align}
  \end{subequations}
According to the Central Limit Theorem $[ \mathbf{e}_l ]_d$ can be approximated as a Gaussian-distributed variable when $D$ is sufficiently large. The parameters of this Gaussian distribution are the marginal mean and variance given in~(\ref{eq:e_moments}).

  The output $\mathbf{c}_l$ at step \ref{eq:l_dense_layer} is then represented as a sum of two Gaussian-distributed vectors: $\mathbf{b}$ and~$\mathbf{e}_l$, i.e. it is a Gaussian-distributed vector with components $c_{d} \sim \mathcal{N}(m^c_{d}, v^c_{d})$, where $m^c_{d} = m^b_{d} + m^e_{d}$ and $v^c_{d} = v^b_{d} + v^e_{d}$.

  Then $\widehat{\boldsymbol\beta}_{l}$ at step \ref{eq:l_thr} of LISTA (Algorithm~\ref{alg:lista}) is the result of soft-thresholding of a Gaussian variable, which is approximated with the spike and slab distribution,  similar to step \ref{eq:thr_first} (section~\ref{subsec:forward_init}).
  Thus, all the steps of BayesLISTA are covered and distributions for outputs of these steps are derived.
  
  \section{Appendix: Backpropagation}
  \label{sec:backpropagation}
  
  The exact intractable posterior (\ref{eq:posterior}) is approximated with a factorised distribution
\begin{align}
\label{eq:approximating_dsitribution}
\begin{split}
q(\mathbf{W}, \mathbf{S}, \gamma, \eta) &= \prod_{d=1}^D\prod_{k=1}^K \mathcal{N}(w_{dk} ; m^w_{dk}, v^w_{dk}) \prod_{d'=1}^D\prod_{d''=1}^D \mathcal{N}(s_{d'd''} ; m^s_{d'd''}, v^s_{d'd''}) \\
&\times \text{Gam}(\gamma; a^\gamma, b^\gamma) \text{Gam}(\eta; a^\eta, b^\eta)
\end{split}
\end{align}
  
  Parameters of approximating distributions are updated with the assumed density filtering (ADF) and expectation propagation (EP) algorithms derived on the derivatives of the logarithm of a normalisation constant (based on~\cite{hernandez2015probabilistic}). ADF iteratively incorporates factors from the true posterior $p$ in (\ref{eq:posterior}) into the factorised approximating distribution $q$ in (\ref{eq:approximating_dsitribution}), whereas EP iteratively replaces factors in $q$ by factors from $p$.
  
  When a factor from $p$ is incorporated into $q$, $q$ has the form $q(a) = Z^{-1}f(a)\mathcal{N}(a; m, v)$ as a function of weights $\mathbf{W}$ and $\mathbf{S}$,
  where $Z$ is the normalisation constant and $f(a)$ is an arbitrary function, $a \in \{w_{dk}, s_{d'd''}\}$. New parameters of the Gaussian distribution for $a$ can be computed as~\cite{minka2001thesis}
  \begin{equation}
  \label{eq:param_update}
  m:= m + v \frac{\partial \log Z}{\partial m}, \,
  v:= v - v^2\left[ \left(\frac{\partial \log Z}{\partial m}\right)^2 - 2 \frac{\partial \log Z}{\partial v}\right]
  \end{equation}
Then for new values of $\mathbf{W}$ and $\mathbf{S}$ derivatives of the logarithm of $Z$ are required when the factor of $p$ is incorporated in $q$.
  
  With the likelihood factors~(\ref{eq:likelihood}) of $p$ the ADF approach is employed and they are iteratively incorporated into $q$. The normalisation constant of $q$ with the likelihood term for the data point $n$ incorporated is (let $z_d$ denote (to simplify notation the superscript $(n)$ is omitted)
  \begin{equation}
  Z  = \int \prod_{d=1}^{D} \mathcal{N}(\beta_d ; [f(\mathbf{y} ; \mathbf{S}, \mathbf{W}, \lambda)]_d, \gamma^{-1}) q(\mathbf{W}, \mathbf{S}, \gamma, \eta) \mathrm{d}\mathbf{W} \mathrm{d}\mathbf{S} \mathrm{d}\gamma \mathrm{d}\eta
  \end{equation}
  Assuming the spike and slab distribution for $\widehat{\boldsymbol\beta}$, the normalisation constant can be approximated as
\begin{equation}
\label{eq:Z}
Z \approx \prod_{d=1}^D \left[\omega^{\widehat{\boldsymbol\beta}}_d  \mathcal{T}\left(\beta_d ; 0, \beta^\gamma / \alpha^\gamma, 2\alpha^\gamma\right) + \vphantom{m^{\widehat{\boldsymbol\beta}}_d} \left(1 - \omega^{\widehat{\boldsymbol\beta}}_d\right)\mathcal{N}\left(\beta_d ; m^{\widehat{\boldsymbol\beta}}_d,  \beta^\gamma / (\alpha^\gamma - 1) + v^{\widehat{\boldsymbol\beta}}_d\right)\right],
\end{equation}
  where $\{\omega^{\widehat{\boldsymbol\beta}}_d, m^{\widehat{\boldsymbol\beta}}_d, v^{\widehat{\boldsymbol\beta}}_d\}$ are the parameters of the spike and slab distribution for $[\widehat{\boldsymbol\beta}]_d$. Parameters of $q$ are then updated with the derivatives of $Z$ according to (\ref{eq:param_update}).
  
  Prior factors (\ref{eq:ws}) and (\ref{eq:gamma_eta}) from $p$ are incorporated into $q$ with the EP algorithm~\cite{hernandez2015probabilistic}, i.e. they replace the corresponding approximating factors from $q$, and then $q$ is updated to minimise the Kullback--Leibler divergence.
  
   \begin{figure}[!t]
  \centering
  \subfloat[MNIST for $K = 100$]{\includegraphics[width=0.25\textwidth]{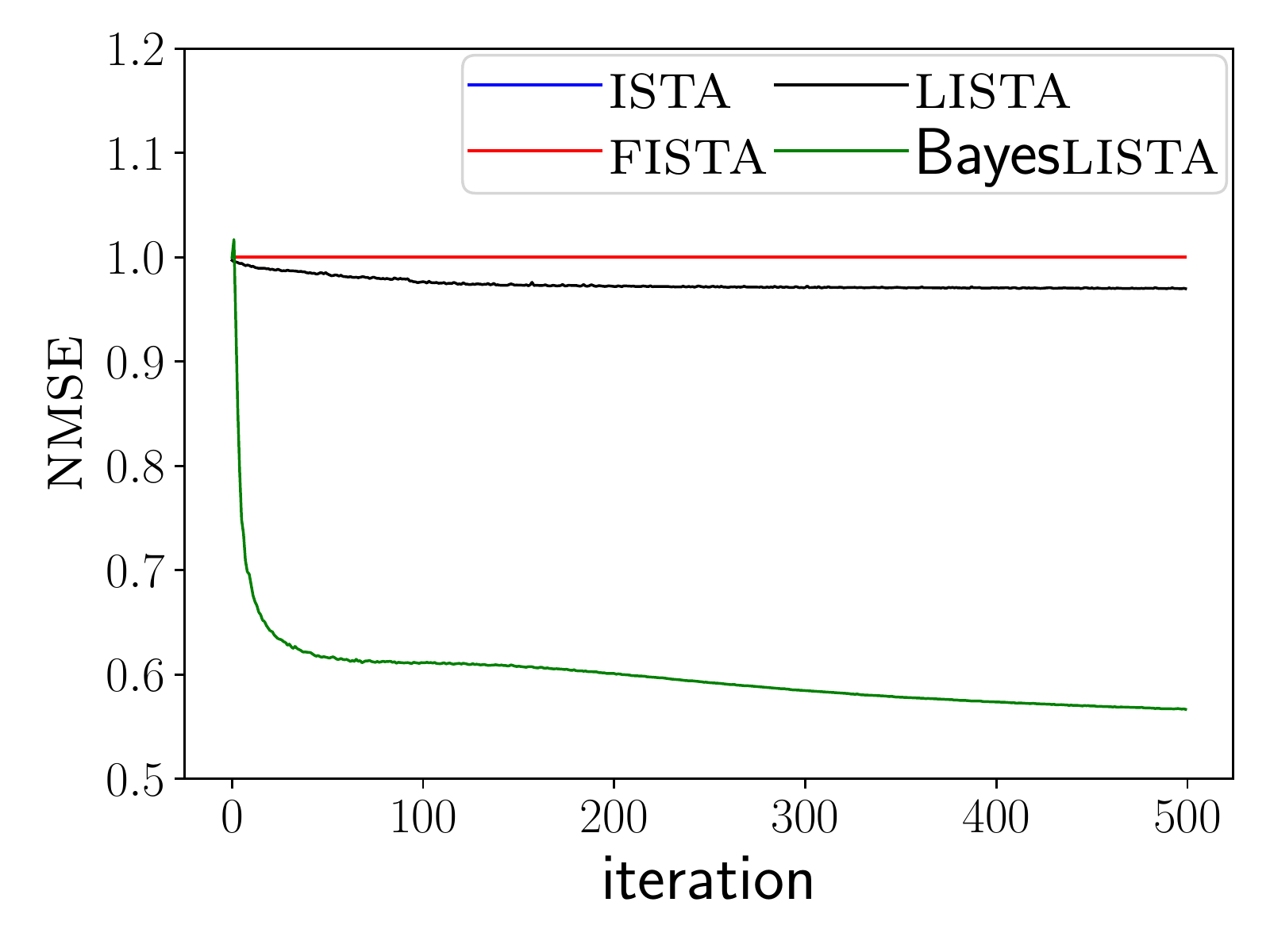}
  \label{fig:nmse_k_100}}
  \hfil
  \subfloat[MNIST for $K = 250$]{\includegraphics[width=0.25\textwidth]{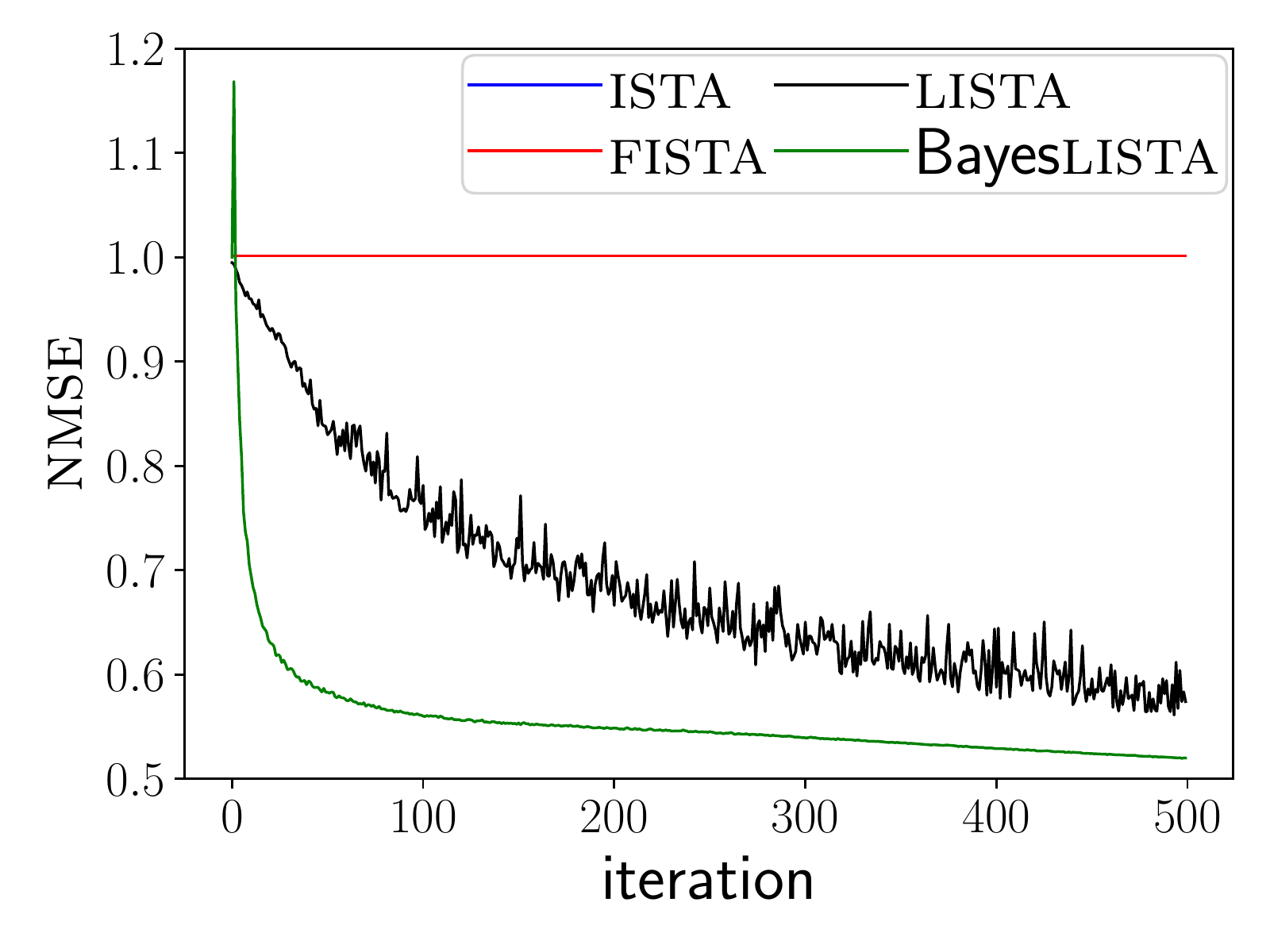}
  \label{fig:nmse_k_250}}
  \caption{NMSE results on the MNIST data for increasing number of iterations with the observation size $K = 100$~\protect\subref{fig:nmse_k_100} and $K = 250$~\protect\subref{fig:nmse_k_250}}
  \label{fig:number_of_layers_synthetic}
  \end{figure}    

\section{Appendix: Predictive performance on MNIST data}
\label{app:mnist}
In this experiment, the methods are evaluated on the MNIST dataset in terms of predictive performance. We use $100$ images for training and $100$ for test.
  
  Figures \ref{fig:nmse_k_100} and~\ref{fig:nmse_k_250} present NMSE with observation sizes $100$ and $250$. The experiment with $K=100$ presents severe conditions for the algorithms: the limited size of the training dataset combined with the small dimensionality of observations. BayesLISTA is able to learn under these conditions, significantly outperforming LISTA. Under better conditions of the second experiment with $K=250$, both NNs converge to the similar results. However, BayesLISTA demonstrates a remarkably better convergence rate. ISTA and FISTA are unable to perform well in these experiments.

\end{document}